\documentclass[10pt,twocolumn,letterpaper]{article}

\usepackage{cvpr}              
\usepackage{graphicx}
\usepackage{amsmath}
\usepackage{amssymb}
\usepackage{booktabs}
\usepackage{adjustbox}
\usepackage{makecell}
\usepackage{array} 
\usepackage{breqn}
\usepackage{xcolor}
\newtheorem{theorem}{Theorem}
\usepackage[normalem]{ulem}
\usepackage{amssymb}
\usepackage{multirow}
\usepackage{multicol}
\usepackage{enumitem}
\usepackage[symbol]{footmisc}

\definecolor{cvprblue}{rgb}{0.21,0.49,0.74}
\usepackage[pagebackref,breaklinks,colorlinks,allcolors=cvprblue]{hyperref}
\usepackage[capitalize]{cleveref}
\crefname{section}{Sec.}{Secs.}
\Crefname{section}{Section}{Sections}
\Crefname{table}{Table}{Tables}
\crefname{table}{Tab.}{Tabs.}
\usepackage[normalem]{ulem}

\usepackage{shortbold}

\def\paperID{2915} 
\def\confName{CVPR}
\def\confYear{2025}

\title{Decision SpikeFormer: Spike-Driven Transformer for Decision Making}
\author{
Wei Huang\textsuperscript{1,2\footnotemark[1]}\\
\textsuperscript{1}Shanghai AI Laboratory\\
{\tt\small huangwe@whu.edu.cn}
\and
Qinying Gu\textsuperscript{1\footnotemark[2]}\\
\textsuperscript{2}Wuhan University\\
{\tt\small guqinying@pjlab.org.cn}
\and  
Nanyang Ye\textsuperscript{3\footnotemark[2]}\\
\textsuperscript{3}Shanghai Jiao Tong University\\
{\tt\small ynylincoln@sjtu.edu.cn}
}

\begin{document}
\maketitle
\footnotetext[1]{This work was done when Wei Huang interned at Shanghai Artificial Intelligence Laboratory.}
\footnotetext[2]{Qinying Gu and Nanyang Ye are co-corresponding authors.}

\begin{abstract}
   Offline reinforcement learning (RL) enables policy training solely on pre-collected data, avoiding direct environment interaction—a crucial benefit for energy-constrained embodied AI applications. Although Artificial Neural Networks (ANN)-based methods perform well in offline RL, their high computational and energy demands motivate exploration of more efficient alternatives. Spiking Neural Networks (SNNs) show promise for such tasks, given their low power consumption. In this work, we introduce DSFormer, the first spike-driven transformer model designed to tackle offline RL via sequence modeling. Unlike existing SNN transformers focused on spatial dimensions for vision tasks, we develop Temporal Spiking Self-Attention (TSSA) and Positional Spiking Self-Attention (PSSA) in DSFormer to capture the temporal and positional dependencies essential for sequence modeling in RL. Additionally, we propose Progressive Threshold-dependent Batch Normalization (PTBN), which combines the benefits of LayerNorm and BatchNorm to preserve temporal dependencies while maintaining the spiking nature of SNNs. Comprehensive results in the D4RL benchmark show DSFormer’s superiority over both SNN and ANN counterparts, achieving 78.4\% energy savings, highlighting DSFormer's advantages not only in energy efficiency but also in competitive performance. Code and models are public at \href{https://wei-nijuan.github.io/DecisionSpikeFormer/}{project page}.
\end{abstract}

\section{Introduction}
\label{sec:intro}
Offline reinforcement learning (RL) aims to develop effective policies solely from pre-collected data that capture agent behaviors without interacting with the environment~\cite{offlineRLtutorial}. This approach is crucial for embodied AI applications, particularly when direct exploration is constrained by safety, energy and resource limitations. While artificial neural networks (ANNs) have driven significant advancements in offline RL—addressing challenges from policy regularization~\cite{kumar2019stabilizing, fujimoto2019off}, value function approximation~\cite{kumar2020conservative, kostrikov2021offline} to conditional sequence modeling (CSM) ~\cite{chen2021decision,janner2021offline}—they often entail substantial computational and energy costs.

Spiking Neural Networks (SNNs), as the third generation of neural networks~\cite{maass1997networks}, offer a promising alternative, bringing low-power, event-driven efficiency to offline RL, especially in energy-constrained, embodied learning scenarios. SNNs have drawn considerable interest for their low energy demands, alignment with biological processes, and compatibility with neuromorphic hardware~\cite{roy2019towards,davies2018loihi,li2024brain}. Unlike ANNs, SNNs operate efficiently by leveraging sparse, event-driven computations triggered only when neurons reach a threshold. This spike-based mechanism allows SNNs to achieve substantial energy savings, making them a powerful choice for neuromorphic applications. SNNs have shown success in vision tasks like image classification and object detection, through CNN-based and more recently, transformer-based architectures~\cite{fang2021deep,deng2022temporal,guo2023rmp,hu2024advancing,li2022spike, zhou2024qkformer,yao2024spikedriven, zhou2023spikingformer, zhou2023spikformer, kim2020spiking,su2023deep,luo2024integer}.

However, applying SNNs to offline RL presents unique challenges. Traditional RL methods rely on precise estimation of state-action values or continuous policy functions, which are challenging to match with SNNs’ discrete spike-based processing. As a result, most approaches focus on ANN-to-SNN conversion or hybrid frameworks combining ANNs and SNNs~\cite{patel2019improved,tan2021strategy}. Recently, offline RL has been approached as a CSM problem~\cite{chen2021decision}, framing it as a sequence prediction task to generate actions with a goal-conditioned policy. Transformers~\cite{vaswani2017attention}, known for their ability to model time-dependent features and manage large-scale sequential data, are widely used to directly model trajectories in offline RL. While spike-driven transformers have proven effective in vision tasks, their application in sequential decision-making tasks with fine-grained temporal dynamics presents additional complexities.

In this work, we introduce Decision SpikeFormer (DSFormer), the first spike-driven transformer model for offline RL. Unlike SNN transformers in vision, which focus on spatial dependencies, DSFormer’s self-attention design emphasizes temporal and positional dependencies. We develop Temporal Spiking Self-Attention (TSSA), which concatenates inputs along the temporal dimension before applying self-attention to enhance global temporal dependencies—essential for effective credit assignment over long sequences. To capture local dependencies and improve energy efficiency, we further introduce Positional Spiking Self-Attention (PSSA) with a learnable positional bias and element-wise computations. The localized window in PSSA reduces computational complexity, aligning well with the efficiency and speed requirements of high-dimensional SNNs. Additionally, we propose Progressive Threshold-dependent Batch Normalization (PTBN) to address the disruptive effects of LayerNorm on SNN spiking behavior. PTBN transitions from LayerNorm to SNN-compatible BatchNorm during training and simplifies to BatchNorm at inference, preserving both temporal dependencies and spiking characteristics.

We summarize our contributions as follows:
\begin{itemize}[itemsep=2pt,topsep=0pt,parsep=0pt]
\item We develop two new self-attention mechanisms, TSSA and PSSA, designed for sequence modeling tasks in offline RL, tailored to capture temporal and positional dependencies, distinguishing them from prior spike-driven self-attention focused on vision tasks. 
\item We introduce PTBN, a progressive normalization method combining the advantages of LayerNorm and BatchNorm that preserve the temporal dependency and maintain the spiking nature of SNNs in the mean time.
\item We validate DSFormer on D4RL, achieving 78.4\% energy savings and surpassing SNN and even ANN counterparts in performance, showing its potential for both energy efficiency and superior task results, marking a significant advancement in SNN applications.
\end{itemize}

\begin{figure*}[ht]
    \centering
    \includegraphics[width=\linewidth]{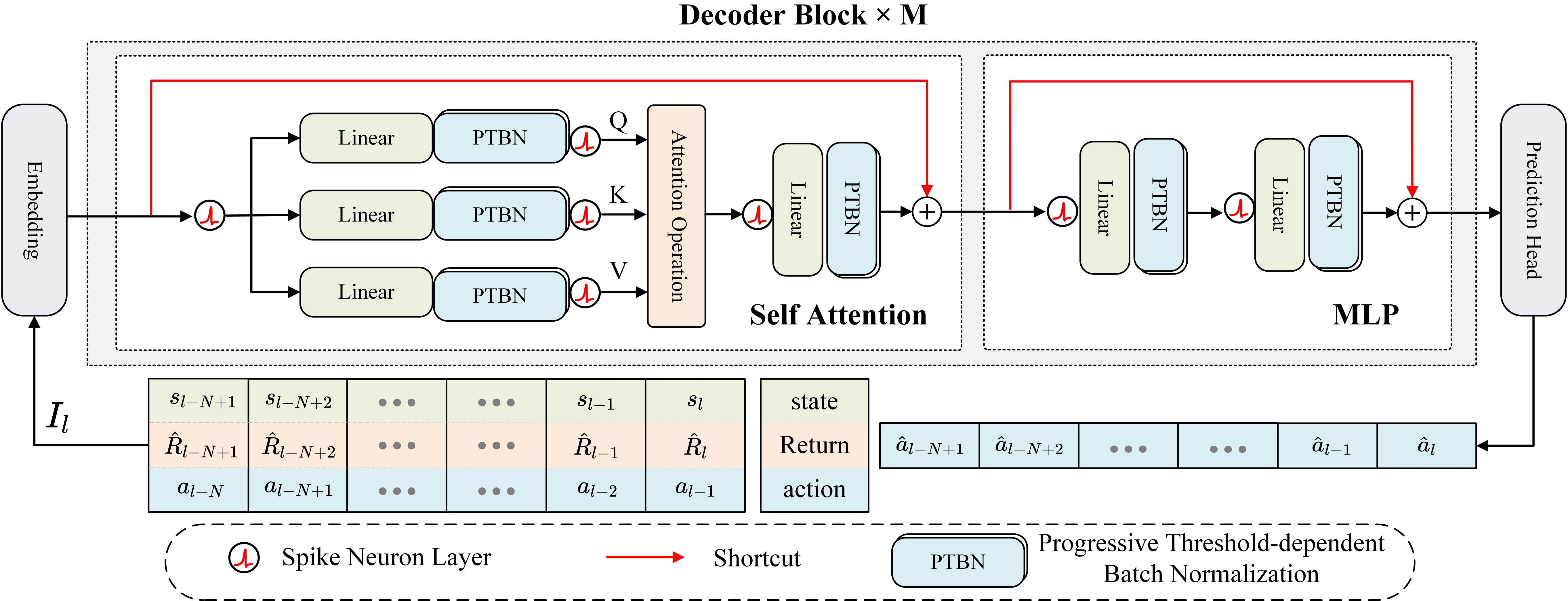}
    \caption{The overall architecture of DSFormer. The input sequence $I_l$ is embedded, repeated $T$ times, fed into the Decoder Blocks through a spike-driven self-attention and MLP layer in each block, and finally passed to the Prediction Head to generate next action predictions.}
    \label{fig:architecture}
\end{figure*}

\section{Related Work}
\subsection{Offline Reinforcement Learning}
In offline RL, early approaches such as behavior cloning (BC)~\cite{torabi2018behavioral} mapped states to actions from demonstrations but were limited by data diversity and distribution mismatch. Policy conservatism and regularization~\cite{hu2023iteratively,kostrikov2021offline,lyu2022mildly,nachum2017bridging} addressed these challenges by constraining learning within the dataset’s support, improving stability and reducing overestimation.

Recent advances have redefined the task as CSM~\cite{emmons2021rvs,janner2021offline,brandfonbrener2022does,hu2023prompt}, frames offline RL as a supervised sequence prediction task. This approach, exemplified by Decision Transformer (DT)~\cite{chen2021decision}, uses transformers to capture long-term dependencies by conditioning actions on past states and future returns, eliminating the need for bootstrapping. The simplicity of the DT structure inspires many follow-up work~\cite{shang2022starformer,brohan2022rt,siebenborn2022crucial,wen2023large} using Transformer~\cite{vaswani2017attention} to model the temporal features across diverse decision-making and control tasks. Recently, Some works pointed out DT's limitation of stitching optimal sequences from suboptimal data, and they address this issue by integrating value-based regularization (e.g. Q-value)~\cite{wang2024critic,hu2024q} to help balance trajectory modeling with optimal action selection. In this work, we focus on designing an SNN model within the DT framework without adding any extra value-based regularization.

\subsection{Spike-driven Transformers}
\noindent\textbf{Vision Tasks} Spikformer~\cite{zhou2023spikformer} pioneered spiking vision transformers, introducing spiking self-attention to eliminate traditional multiplications and softmax, using spiking neurons and BatchNorm in place of LayerNorm and GELU. Spikingformer~\cite{zhou2023spikingformer} proposed a spike-driven residual learning to avoid non-spike computations in Spikformer. SpikeFormer~\cite{yao2023spikedriven} and Meta-SpikeFormer~\cite{yao2024spikedriven} developed four spike-driven self attentions, making SNN transformers applicable across tasks like classification and segmentation. SpikingResformer~\cite{shi2024spikingresformer} also showed competitive vision task results with a ResNet-inspired, dual self-attention design. \\
\noindent\textbf{Sequence Modeling Tasks} Spike-driven transformers have also advanced in sequence modeling for NLP tasks. SpikeGPT~\cite{zhu2024spikegpt}, for example, replaces standard self-attention with an element-wise spiking RWKV module to maintain temporal dependencies, while SpikeBERT~\cite{lvspikebert} directly adapts Spikformer to the BERT architecture. Similarly, SNN-BERT~\cite{su2024snn} introduces bidirectional spiking neurons to enhance temporal modeling capabilities in NLP. However, despite these advances, many of these models still rely on floating-point calculations, particularly in softmax and LayerNorm, during training and inference. More detailed Related Work is presented in \textbf{Appendix B}.

\section{Method}
\subsection{Preliminary}
\paragraph{Spiking Neural Network} In the spiking neuron layer, we adopt the Leaky Integrate-and-Fire (LIF) model, known for its ability to mimic biological neuron behavior and efficiently handle temporal information~\cite{maass1997networks}. The dynamics of the LIF neuron are described by the following equation:

\begin{equation}
\begin{aligned}
&\begin{gathered}
U^t=H^{t-1}+I^t \\
S^t=\operatorname{Hea}\left(U^t-U_{t h}\right)
\end{gathered}\\
&H^t=U_{\text {reset }} S^t+\gamma U^t\left(1-S^t\right)
\end{aligned}
\end{equation}
where $I^t$ is the input to the LIF neuron at time step $t$, $U^t$ is the membrane potential that integrate $I^t$ and temporal input $H^{t-1}$. $\mathrm{Hea}(\cdot)$ is a Heaviside function which equals 1 for $x \geq 0$ and 0 otherwise. When $U^t$ exceeds the firing threshold $U_{t h}$, the neuron fires a spike signal $S^t$ and the membrane potential is reset to $U_{\text {reset }}$; otherwise, the membrane potential decays to $H^t= \gamma U^t$, where $\gamma \leq 1$ is the decay factor of the membrane potential. For simplicity, we denotes the spiking neuron layer as $\Scal\Ncal(\cdot)$ in subsequent sections.

\paragraph{Offline RL}

The goal of RL is to learn a policy maximizing the expected cumulative return $\mathbb{E}[R(\tau)]$ in a Markov Decision Process (MDP) which is formulated by the tuple $(\mathcal{S}, \mathcal{A}, \mathcal{P}, \mathcal{R})$, including states $s \in \mathcal{S}$, actions $a \in \mathcal{A}$, transition dynamics $P(s'|s,a)$, and a reward function $r = \mathcal{R}(s,a)$, where $\tau$ denotes the trajectory sequence. In offline RL, algorithms learns a policy $\pi_{\theta}$ entirely from a static dataset $\mathcal{D} = \{\tau|\tau \sim \pi_{\beta},\tau = (s_0, a_0, r_0, \dots, s_{L}, a_{L}, r_{L})$ collected by an unknown behavior policy $\pi_{\beta}$. To generate actions based on future desired rewards, we adopt the method to feed return-to-go (i.e., the suffix of the reward sequence, $\widehat{R}_{l}=\sum_{l^{\prime}=l}^{L}r_{l}$) into the trajectory $\tau$ instead of feeding the rewards directly~\cite{chen2021decision}.

\subsection{Overall Architecture}
The DSFormer, inspired by the DT, employs an autoregressive Transformer architecture, as illustrated in Figure~\ref{fig:architecture}. The network consists of an Embedding Layer, $M$ stacked Decoder Blocks, and a Prediction Head. Each Decoder Block follows standard SNN Transformer designs~\cite{zhou2023spikformer, yao2024spikedriven}, incorporating spike-driven Self-Attention and a spike-driven Multi-Layer Perceptron (MLP). In the Self-Attention layer, two spiking self-attention mechanisms, TSSA and PSSA are proposed considering the Markov nature in offline RL. PTBN is proposed and used instead of BatchNorm to accommodate the autoregressive sequence decision-making process in SNNs, and Membrane Shortcut (MS)~\cite{hu2024advancing} is adopted in the achitecture. Both the Embedding Layer and the Prediction Head are implemented through fully connected layers.

At each iteration step $l$ for the agent to interact with the environment, the input sequence is formalized as: $I_l=(a_{l-N}, \hat{R}_{l-N+1}, s_{l-N+1}, \ldots, a_{l-1}, \hat{R}_l, s_l)$, where $N$ represents the context length of historical information for making a decision. $a$, $\hat{R}$, and $s$ represent the action, estimated return-to-go, and state, respectively. Note that for clarity, $t$ denotes the inner time step of the SNN, while $l$ refers to the iteration step in offline RL. Each element in $I_l$ is concatenated and embedded to form the input matrix as:
 $X_l= \{ \operatorname{Emb}(a_{l-N}, \hat{R}_{l-N+1}, s_{l-N+1}); \operatorname{Emb}(a_{l-N+1}, \hat{R}_{l-N+2}, \\ s_{l-N+2}); \ldots; \operatorname{Emb}(a_{l-1}, \hat{R}_l, s_l) \} \in \mathbb{R}^{N \times D}$, where $D$ is the channel dimension, and the number of tokens equals the context length $N$. $X_l$ is then repeated $T$ times along the temporal dimension (T being the total SNN timesteps) and fed into the Decoder Blocks. For a comprehensive notation table, see \textbf{Appendix A}.

In each Decoder Block, the input $X$ passes through a spike-driven Self-Attention layer to yield $Y=X+\operatorname{Self-Attention}\left(X \right)$, followed by an MLP layer to produce the final output $Z=Y+\operatorname{MLP}\left(Y \right)$. The output from the last Decoder Block is normalized along the temporal dimension and passed to the Prediction Head to generates the next action predictions.

\begin{figure*}[ht]
    \centering
    \includegraphics[width=\linewidth]{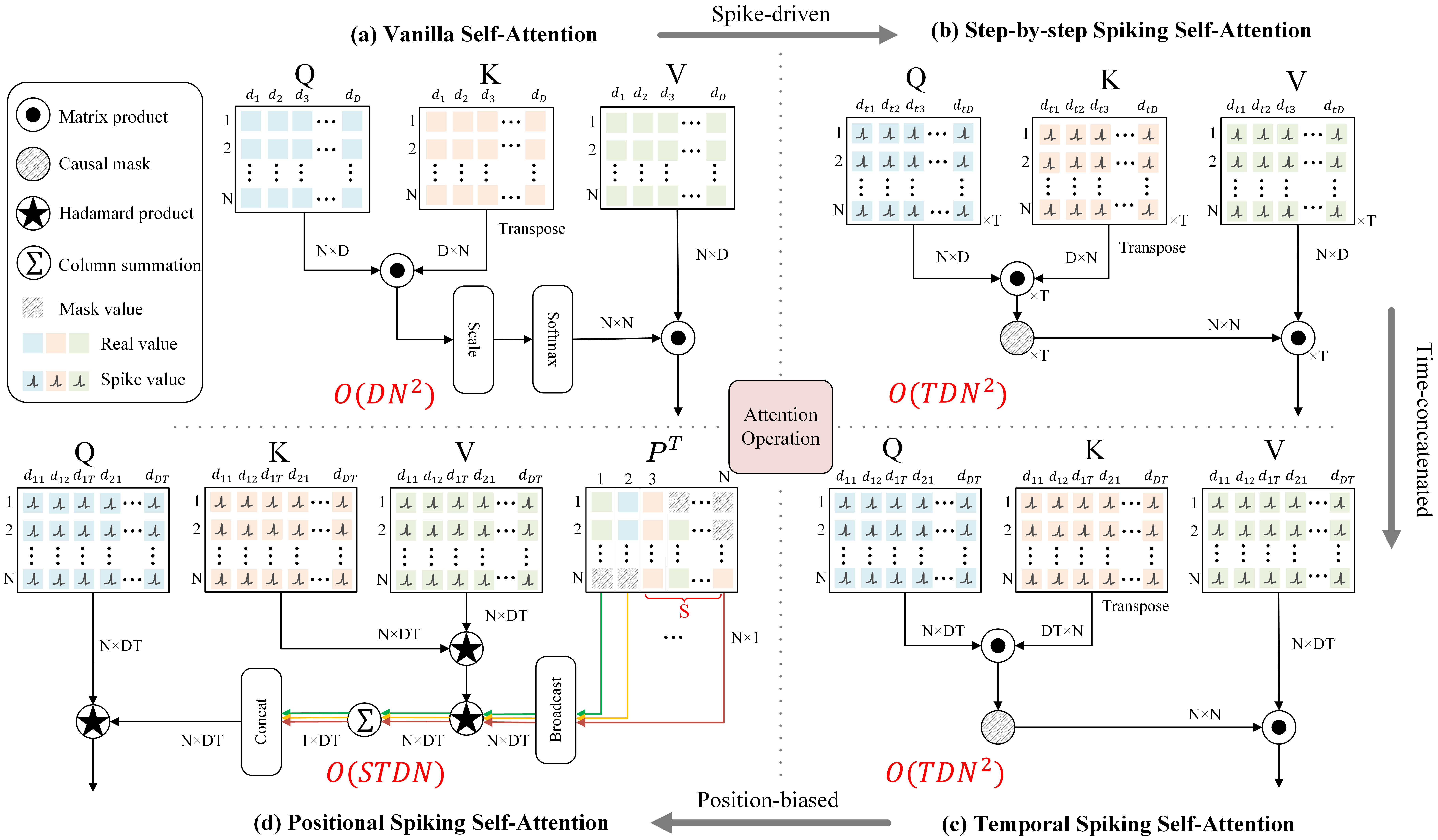}
    \caption{Self-attention mechanisms with different computational complexity. (a) VSA inherited from~\cite{vaswani2017attention}. (b) SSSA, a spike version of VSA. (c) TSSA that concatenates inputs across the temporal dimension before self-attention. (d) PSSA that incorporates positional bias. To simplify the plotting, we set T = 3.}
    \label{fig:self-attention}
\end{figure*}

\subsection{Spike-driven Self-attention}

\noindent\textbf{Vanilla Self-Attention}

As shown in Figure~\ref{fig:self-attention}(a), the vanilla self-attention (VLA) can be formulated as follows. For clarity, we will explain within a single attention head in the following discussions:
\begin{equation}
\begin{aligned}
& Q ={X W_q}, {K}={X} {W_k}, {V}={X W_v} \\
& \operatorname{Attention}({Q}, {K}, {V})=\operatorname{Softmax}\left(\frac{{Q K}^{{\top}}}{\sqrt{d_{k}}}\right) {V}
\end{aligned}
\end{equation}
where $Q$, $K$ and $V$ are the query, key, and value matrices, obtained by applying three learnable linear transformations to the original input ${X}$, with ${W}_q$, ${W}_k$ and ${W}_v$ as the corresponding weight matrices. $d_k$ is the channel dimension of $K$. The VLA mechanism has a high quadratic complexity of $O(DN^2)$, with respect to the token numbers. Besides, the Softmax and division operations are not compatible with the addition nature of SNNs.

% \\ \hspace*{\fill} \\
\noindent\textbf{Step-by-step Spiking Self-Attention}
To address this, we first propose the Step-by-Step Spiking Self-Attention (SSSA) shown in Figure~\ref{fig:self-attention}(b),
which is inspired by the design of the spiking self attention (SSA)~\cite{zhou2023spikformer} but with a causal masking. Intuitively, the causal masking prevents predicting the current action with the next state, action, and return-to-go tuple. SSSA repeats self-attention computations at each time step. As shown in Equation~\ref{eq:SSSA}, the spike matrices $Q,K,V$ are first generated by mapping the input $X$ with three weight matrices and then normalizing them with the proposed PTBN, which will be explained later. 
\begin{align}
Q, K, V = & \, \Scal\Ncal \left(\operatorname{PTBN}\left(X W_q\right)\right), 
        \Scal\Ncal \left(\operatorname{PTBN}\left(X W_k\right)\right), \nonumber \\
          & \, \Scal\Ncal \left(\operatorname{PTBN}\left(X W_v\right)\right)
\label{eq:SSSA}
\end{align}
where $W_q$, $W_k$ and $W_v$ are the linear transformation matrices for $Q$, $K$ and $V$, and $\Scal\Ncal(\cdot)$ is the spiking neuron layer. The SSSA computation at each time step is given by:
\begin{equation}
\operatorname{Attention}\left(Q^t, K^t, V^t\right)=\operatorname{mask}\left(Q^t (K^t)^\top\right) V^t
\end{equation}
where $Q^t$, $K^t$ and $V^t$ correspond to the input of $Q$, $K$ and $V$ at $t$-th time step, respectively. A causal mask is applied to prevent the model from attending to future information. The overall time complexity for SSSA is $O(TDN^2)$, with its calculation presented in \textbf{Appendix C}

% \\ \hspace*{\fill} \\
\noindent\textbf{Temporal Spiking Self-Attention}
As shown in Figure~\ref{fig:self-attention} (c), Temporal Spiking Self-Attention (TSSA) is then introduced to address the limitation of SSSA. Due to its single-step property, SSSA cannot effectively exploit the temporal nature of SNNs to better capture dependencies within the (actions, states, rewards-to-go) tuples. TSSA overcomes this problem by concatenating the input across the temporal dimension before performing self-attention. This allows the model to weigh the dependencies over the entire sequence simultaneously. The TSSA computation is formulated as: 
\begin{equation}
\operatorname{Attention}(Q, K, V)=\operatorname{mask}\left(Q K^\top\right) V
\end{equation}
where $Q,K,V$ are concatenated $Q^{t},K^{t}, V^{t}$ respectively. This distinguishes the TSSA from previous spike-driven attentions used for image classification tasks, which do not involve complex temporal structures.
We further prove that the temporal-concatenated inputs can capture essential information than the non-concatenated version by reducing the entropy. We prove this from an information-theoretic perspective:

\begin{theorem}
Let $X^t$ represent the input at time step $t$, then the joint entropy after concatenation satisfies: $H\left(X^1, X^2, \ldots, X^T\right)=H\left(X^1\right)+H\left(X^2 \mid X^1\right)+\ldots H\left(X^T \mid X^1, X^2, \ldots, X^{T-1}\right)$. Since 
$X^t$ and $X^{t+1}$ are not independent according to the LIF dynamics: $H\left(X^{t+1} \mid X^1, X^2, \ldots, X^t\right)<H\left(X^{t+1}\right)$. Therefore, we can conclude: $H\left(X^1, X^2, \ldots, X^T\right)<H\left(X^1\right)+H\left(X^2\right)+\ldots+H\left(X^T\right)$.
\end{theorem}
\textbf{Remark:} The Theorem indicates that TSSA reduces input entropy, facilitating more effective pattern learning. Its time complexity remains $O(TDN^2)$, 

comparable to SSSA, without added computational cost.

% \\ \hspace*{\fill} \\
\noindent\textbf{Positional Spiking Self-Attention} While TSSA enhances the modeling of temporal properties in offline RL, it has limitations: (1) it lacks local-dependency modeling and (2) has a quadratic complexity with respect to the token number $N$. To address these issues, motivated by \cite{zhai2021atf,song2024linearsnn}, we propose Positional Spiking Self-Attention (PSSA), which captures local dependencies more effectively in offline RL. Unlike SSSA and TSSA that implement spiking versions of VLA without considering the locality inherent to RL trajectories modeled as Markov decision processes (MDP), PSSA introduces a learnable pair-wise positional bias to model these local associations. This design not only improves modeling accuracy but also enhances speed and reduces energy consumption.
As illustrated in Figure~\ref{fig:self-attention} (d), PSSA incorporates spiking mechanisms, temporal concatenation, and learnable pair-wise positional biases. The PSSA computation is:
\begin{equation}
\operatorname{Attention}(Q, K, V)_i=Q_i \odot \sum_{j=1}^N P_{i j} \odot K_j \odot V_j
\end{equation}
where $\odot$ denotes element-wise multiplication, and $Q_i$, $K_j$ and $V_j$ represent the feature vectors corresponding to the $i$-th query and $j$-th key-value tokens, respectively. The matrix $P \in \mathbb{R}^{N \times N}$ consists of learnable pair-wise positional biases, where each element $P_{i j}$ captures the positional relationship between the $i$-th query token and the $j$-th key-value token pair. This formulation first combines keys and values to capture token-to-token dependencies, then modulates these interactions with positional biases, and finally integrates with the query vector to produce context-aware information for each token. To account for the local dependencies of offline RL tasks, we use the following learnable relative positional bias:
\begin{equation}
P_{i j}= \begin{cases}P_{i j}, & \text { if }|i-j|<S \\ 0, & \text { otherwise. }\end{cases}
\end{equation}
where $S$ is the local window size.

The time complexity of PSSA is $O(STDN)$; since $S$ is a constant with $S \ll N, D$, the time complexity can be approximated as $O(TDN)$.

\subsection{Progressive Threshold-dependent Batch Normalization (PTBN)}
\begin{figure}[t]
\centering
\centerline{\includegraphics[width=0.95\columnwidth]{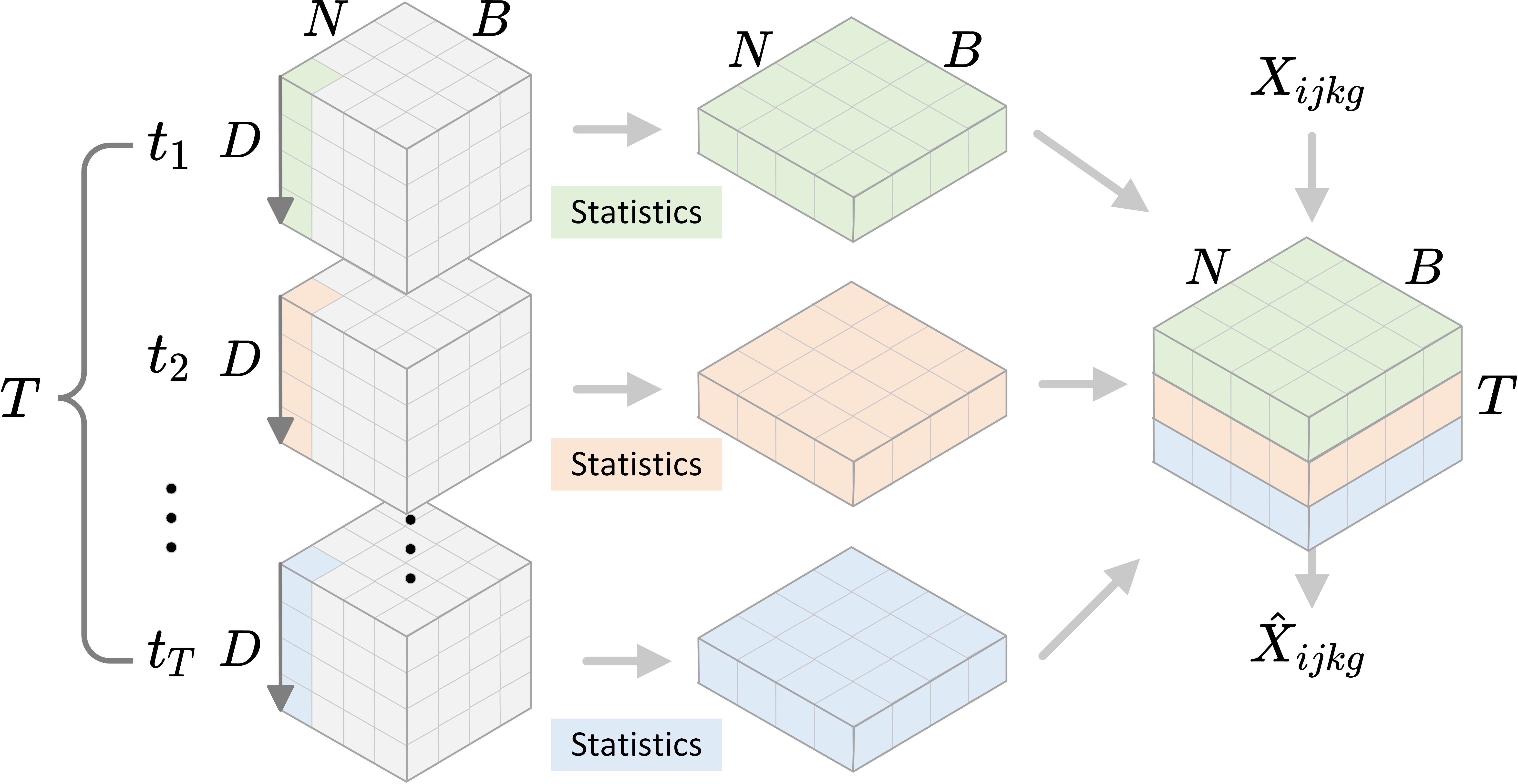}}
\caption{Design of tdLN. Each cube represents the feature map at timestep \( t \) and calculates statistics along the \( D \) dimension to obtain mean and variance with a shape of \( B \times N \times T \) for normalization.}
\vspace{-1em}
\label{fig:tdLN}
\end{figure}

Threshold-dependent batch normalization (tdBN)~\cite{zheng2021going} is widely used and essential in deep SNNs to mitigate gradient explosion and vanishing. However, in sequence modeling for offline RL, preserving dependencies within each sequence is essential, as each iteration step relies on context from surrounding steps. tdBN, which normalizes across the batch dimension, may disrupt these dependencies by introducing unintended interactions between sequences. In contrast, LayerNorm is more appropriate as it can normalize each time and iteration step independently, preserving temporal dependencies and ensuring stability across variable-length sequences. The combination of Linear layer and LayerNorm is widely adopted in standard Transformer where LayerNorm normalizes inputs along the feature dimension without accounting for the sequence length. To adapt this for SNN sequence modeling, we first propose Threshold-dependent Layer Normalization (tdLN) as shown in Figure~\ref{fig:tdLN}. 

Specifically, the input of the tdLN is denoted as  $X \in \mathbb{R}^{B \times N \times T \times D}$, where $B$ is the batchsize. A threshold voltage $U_{\mathrm{th}}$ and a scaling hyperparameter $\alpha$ are introduced to control the magnitude of normalized values, making them compatible with the threshold mechanism in SNNs. The tdLN is performed as:

\begin{equation}
\hat{X}_{i j k g}=\frac{\alpha U_{\mathrm{th}}\left(X_{i j k g}-\mu_{ijk}\right)}{\sqrt{\sigma_{ijk}^2+\epsilon}}, \quad Y_{i j k g}=\lambda \hat{X}_{i j k g}+\beta
\end{equation}
where $\epsilon$ is a small positive constant to prevent division by zero, $\mu_{ijk}$ and $\sigma_{ijk}^2$ are the estimated mean and variance of $X$ across the channel dimension $D$. We do not normalize over the context dimension $N$ as in offline RL, the state should change drastically due to actions. $\lambda$ and $\beta$ are learnable parameters.

However, since $\mu_{ijk}$ and $\sigma_{ijk}^2$ vary for each batch of input (i.e. $\mu_{ijk}, \sigma_{ijk}^2 \in \mathbb{R}^{B \times N \times T}$), they are incompatible with the fixed weights $W \in \mathbb{R}^{D \times D'}$ and biases $b \in \mathbb{R}^{D'}$ of the Linear layer. This prevents merging the Linear layer and tdLN during inference, resulting in additional floating-point operations that disrupt the spiking nature of SNNs. To resolve this, motivated by \cite{guo2024slab}, we propose Progressive Threshold-dependent Batch Normalization (PTBN), which gradually transitions from tdLN to a SNN-compatible tdBN during training. According to \cite{zheng2021going}, tdBN here is performed as $\frac{\alpha U_{\mathrm{th}}\left(X_{i j k g}-\mu_{g}\right)}{\sqrt{\sigma_{g}^2+\epsilon}}$. PTBN combines tdLN and tdBN using a gradually updated weight:
\begin{equation}
\begin{aligned}
\operatorname{PTBN}(x) & =\theta \operatorname{tdLN}(x)+(1-\theta) \operatorname{tdBN}(x)
\end{aligned}
\end{equation}
The parameter $\theta$ controls the balance between tdLN and the tdBN, fine-tuning the adjustment during training. Initially, $\theta$ starts at 1, favoring tdLN, and linearly decreases to tdBN as training progresses:
\begin{equation}
\theta=\frac{T_p-T_{c u r}}{T_p}, \theta \in[0,1]
\end{equation}
where $T_p$ is the predetermined training step for PTBN, and 
$T_{c u r}$ is the current training step.

During training, $T_p$ is allocated a predefined proportion of total steps, with the remainder dedicated to tdBN for BatchNorm adaptation.
During inference, PTBN reduces to the tdBN form, which can be merged with fully connected layers, eliminating extra computations and preserving the spiking characteristics of SNNs. Further details on the calculation and implementation can be found in the \textbf{Appendix D}.

\begin{table*}[!h]
\centering
\caption{\textbf{Results on MuJoCo.} We report mean and variance scores from five random seeds, following \cite{fu2020d4rl}. Dataset abbreviations: `medium' as `m' (medium policy achieving one-third of expert score); `medium-replay' as `m-r' (mixed-quality trajectories from training); `medium-expert' as `m-e' (trajectories combining medium and expert policies). The strengthened digits denote the highest scores. }
\resizebox{\textwidth}{!}{
\setlength{\tabcolsep}{3mm}{
\begin{tabular}{c||cccc|cc|cc}  
\hline\hline
\textbf{MuJoCo Tasks}& \textbf{BC} &\textbf{CQL} &\textbf{DT} &\textbf{FCNet} &\textbf{SpikeGPT} &\textbf{SpikeBert} &\textbf{TSSA} &\textbf{PSSA}\\
\hline
halfcheetah-m-e   &   35.8      &62.4        &86.8$\pm$1.3    &91.2$\pm$0.3   &23.6$\pm$4.5              &24.3$\pm$6.0                & \underline{91.3$\pm$0.2}  & \textbf{91.5$\pm$0.3} \\

walker2d-m-e       &   6.4       &98.7         &108.1$\pm$0.2   &108.8$\pm$0.1  &22.6$\pm$4.8              &92.5$\pm$22.4
& \underline{108.6$\pm$0.2} & \textbf{108.9$\pm$0.1} \\

hopper-m-e        &   \textbf{111.9}     &\underline{111.0}        &107.6$\pm$1.8   &110.5$\pm$0.5  &32.7$\pm$5.4            &84.1$\pm$8.8
&\underline{111.0$\pm$0.7} &110.9$\pm$0.2 \\
 
halfcheetah-m    &   36.1      &44.4         &42.6$\pm$0.1    &\textbf{42.9$\pm$0.4}   &26.9$\pm$0.8            &20.0$\pm$3.5
&42.5$\pm$0.4  &\underline{42.8$\pm$0.3}  \\

walker2d-m        &   6.6       &79.2         &74.0$\pm$1.4    &\textbf{75.2$\pm$0.5}   &16.4$\pm$10.2           &22.9$\pm$10.4
&\underline{72.4$\pm$4.5}  &\textbf{75.2$\pm$1.4}   \\

hopper-m          &29.0         &58.0         &\underline{67.6$\pm$1.0}    &57.8$\pm$6.0   &25.1$\pm$6.4           &31.4$\pm$4.9
&64.6$\pm$2.1  &\textbf{74.1$\pm$4.3}  \\

halfcheetah-m-r   &38.4         &46.2         &36.6 $\pm$0.8  &\textbf{39.8$\pm$0.8}   &21.8$\pm$2.0           &32.2$\pm$8.0
&38.7$\pm$1.1 &\underline{38.8$\pm$0.7}\\

walker2d-m-r    &11.3          &26.7          &\underline{66.6$\pm$3.0} &63.5$\pm$7.5   &16.7$\pm$3.3           &21.2$\pm$6.4
&66.0$\pm$3.3  &\textbf{71.0$\pm$3.6} \\

hopper-m-r      &11.8  &48.6    &82.7$\pm${7.0} 
&\underline{85.8$\pm$1.7}   &51.5$\pm$7.1           &30.1$\pm$8.6
&\underline{85.8$\pm$3.6} &\textbf{96.3$\pm$1.3} \\
\hline
\textbf{MuJoCo mean $\uparrow$} &31.9 &63.9 &74.7 &75.1 &26.4 &39.9 &\underline{75.7} &\textbf{78.8}\\
\hline
\textbf{Power ($\mu$J) $\downarrow$ } & N/A & N/A & 410.5 & 1022.03 & \textbf{27.8} & 806.7 &96.1 & \underline{88.8}\\
\hline\hline
\end{tabular}}
}
\label{tab:Mujoco}
\end{table*}

\begin{table*}
\centering
\caption{\textbf{Results on Adroit.} Mean and variance scores from five random seeds, following \cite{fu2020d4rl}. Abbreviations: `expert' as `m-e' (expert policy); `human' as `h' (human demonstrations); and `cloned' as `c' (behavior cloning). The strengthened digits denote the highest scores.} 
\resizebox{\textwidth}{!}{
\setlength{\tabcolsep}{3mm}{
\begin{tabular}{c|cccc|cc|cc}  
\hline\hline
\textbf{Adroit Tasks}& \textbf{BC} &\textbf{CQL} &\textbf{DT} &\textbf{FCNet} &\textbf{SpikeGPT} &\textbf{SpikeBert} &\textbf{TSSA} &\textbf{PSSA}\\
\hline
pen-e   &   85.1      &107.0           &\underline{110.4$\pm$20.9}    &108.0$\pm$11.3  &30.5$\pm$10.3        &46.2$\pm$19.5
&104.6$\pm$13.2  &\textbf{122.0$\pm$17.8} \\

door-e       &   34.9       &101.5        &95.5$\pm$5.7  &102.9$\pm$2.9 &65.3$\pm$16.9        &96.4$\pm$4.6
&\underline{105.0$\pm$0.3} &\textbf{105.2$\pm$0.1} \\

hammer-e       &   125.6     &86.7        &89.7$\pm$24.6   &121.1$\pm$6.1  &51.1$\pm$18.7        &71.3$\pm$16.5
&\underline{126.4$\pm$0.4} &\textbf{127.2$\pm$0.3} \\

relocate-e    &   101.3      &95.0        &15.3$\pm$3.6    &50.0$\pm$6.0   &0.7$\pm$0.9          &0.3$\pm$0.5
&\underline{106.3$\pm$2.6} &\textbf{108.4$\pm$2.2}  \\

pen-h        &   34.4      &37.5          &-0.2$\pm$1.8   &57.7$\pm$11.1   &29.8$\pm$11.7          &20.0$\pm$16.7  
&\textbf{89.7$\pm$10.0}  &\underline{75.7$\pm$25.1}   \\

door-h       &\underline{0.5}         &\textbf{9.9}    &0.1$\pm$0.0   &0.4$\pm$0.5   &0.1$\pm$0.0          &0.2$\pm$0.0 
&0.4$\pm$0.1  &0.2$\pm$0.0  \\

hammer-h  &\underline{1.5}         &\textbf{4.4}        &0.3$\pm$0.0  
&1.2$\pm$0.0   &0.3$\pm$0.0          &0.3$\pm$0.0
&0.4$\pm$0.1 &0.2$\pm$0.0\\
 
relocate-h    &0.0 &0.2     &\textbf{0.2$\pm$0.2} 
 &0.0$\pm$0.0  &\underline{0.1$\pm$0.0}          &0.0$\pm$0.0
 &0.0$\pm$0.0  &0.0$\pm$0.0 \\

pen-c     &\textbf{56.9}  &39.2    &22.7$\pm$17.1 
&50.4$\pm$24.1 &17.0$\pm$22.0          &17.6$\pm$29.0
&41.1$\pm$19.7 &\underline{44.8$\pm$14.7} \\

door-c    &-0.1  &\textbf{0.4}      &0.1$\pm$0.0   
&-0.2$\pm$0.0  &\underline{0.2$\pm$0.0}          &\underline{0.2$\pm$0.0}   
&\textbf{0.4$\pm$0.8} &0.0$\pm$0.0 \\

hammer-c  &\underline{0.8}   &\textbf{2.1}      &0.3$\pm$0.0   
&0.2$\pm$0.0   &0.3$\pm$0.0          &0.3$\pm$0.5   
&0.2$\pm$0.0 &0.2$\pm$0.0\\

relocate-c &-0.1 &-0.1     &-0.3$\pm$0.0  
&-0.2$\pm$0.0  &\textbf{0.1$\pm$0.0}          &\underline{0.0$\pm$0.5}  
&-0.2$\pm$0.0 &-0.2$\pm$0.0\\
\hline
\textbf{Adroit Mean $\uparrow$} &36.7 &40.3 &27.8 &41.0 &16.3 &21.1 &\underline{47.9} &\textbf{48.6}\\
\hline\hline
\end{tabular}}
}
\label{tab:Adroit}
\end{table*}

\section{Experiments}
\subsection{Experiment Setting}
In this section, we present a comprehensive evaluation of DSFormer using the widely recognized D4RL benchmark~\cite{fu2020d4rl}. A broad range of ANN and SNN baselines are included. Ablation studies are also conducted to examine the individual contributions of key components. We further discuss why DSFormer’s SNN structure is well-suited for offline RL tasks, emphasizing its advantages in handling temporal dependencies and energy-efficient learning.

\paragraph{Baseline}
We begin with classic MLP-based and transformer-based ANN methods, including Behavior Cloning (BC)~\cite{torabi2018behavioral}, Conservative Q-Learning (CQL)~\cite{kumar2020conservative}, Decision Transformer (DT)~\cite{chen2021decision}, and Fourier Controller Network (FCNet)~\cite{fcnet}. 

For SNN architectures, since no specific SNN models are designed for offline RL, we select SNN models from NLP, including SpikeGPT~\cite{zhu2024spikegpt} and SpikeBERT~\cite{lvspikebert}, considering the sequence modeling nature of our work.

\paragraph{Datasets and settings}
We evaluate the DSFormer on the MuJoCo and Adroit environments, which encompass a variety of complex control tasks. The MuJoCo environments—such as HalfCheetah, Hopper, and Walker2d—cover various locomotion modes with continuous action spaces and dense rewards. The Adroit datasets, based on noisy human demonstrations, include tasks like Door, Hammer, and Relocate, which emphasize precise manipulation in high-dimensional spaces. All models are trained on these collected datasets and evaluated in simulators. We report normalized scores following the protocol in \cite{fu2020d4rl}, where a score of 100 represents expert-level performance, and a score of 0 reflects random agent performance.

\paragraph{Number of parameters}
To ensure a fair comparison between DSFormer and its ANN counterpart, DT, both models are based on the same MetaFormer~\cite{yu2023metaformer} architecture, sharing almost identical parameter counts. We apply PTBN before each spiking neuron in DSFormer to prevent gradient explosion during training. While PTBN adds extra parameters during training, it is merged into the Linear layer at inference, resulting in nearly the same parameter counts for DSFomer and DT (0.3\% fewer for TSSA and 0.08\% more for PSSA, compared to DT).

We set $T=4$ for all tasks. Further details on experimental and implementation settings are 
provided in \textbf{Appendix E}.

\subsection{Experiment results}
\paragraph{Results for MuJoCo Domain} 
The experimental results on the MuJoCo environment (Table~\ref{tab:Mujoco}) demonstrate the superior performance of DSFormer. Both spike attention mechanisms, TSSA and PSSA, not only outperform other transformer-based SNN methods by a significant margin but also achieve top average scores (75.7 and 78.8, respectively) in the ANN domain, with substantially improved energy efficiency (410.5 $\mu$J in DT versus 96.1 $\mu$J in TSSA and 88.8 $\mu$J in PSSA). 

SNN architectures perform poorly because: SpikeBERT's shortcut connection introduces integer values that disrupt residual propagation, compromising spike characteristics, while SpikeGPT, designed for NLP, struggles with the limited data in offline RL. 
TSSA and PSSA, however, consistently rank first or second across all sub-tasks. 

Additionally, PSSA excels in most tasks, underscoring its ability to efficiently capture local dependencies in the motion-control MDP due to the introduction of pair-wise positional bias. Further analysis is provided in the Ablation study. Moreover, PSSA’s element-wise multiplication and small local window size ($S=8$) yield lower energy consumption (31.0 $\mu$J) in Self-Attention mechanism than TSSA (44.6 $\mu$J) and DT (168.2 $\mu$J), as detailed in \textbf{Appendix F}.

\paragraph{Results for Adroit Domain} Table~\ref{tab:Adroit} highlights the effectiveness of DSFormer, which significantly outperforms other ANN and SNN methods. Both TSSA and PSSA surpass DT across all sub-tasks, with an substantial performance gains (47.9 and 48.6 versus 27.8). DSFormer’s advantages are more pronounced in Adroit compared to MuJoCo, as its temporal dynamics effectively capture the long-range, high-dimensional action dependencies and sparse rewards essential for precise manipulation tasks in Adroit, as discussed further in Section 4.4. Overall, DSFormer shows exceptional capability in managing complex motion patterns in an energy-efficient, low-power manner. Visualization results are presented in \textbf{Appendix Figure. S1}.

\begin{table}[t]
\centering
\caption{Ablation Study on DSFormer}
\begin{adjustbox}{width=\columnwidth}
\begin{tabular}{cccccc}  
\toprule
Ablation & Methods & Hopper-m-e & Hopper-m & Hopper-m-r & Average \\
\midrule
\multirow{2}{*}{\textbf{Ours}} & TSSA & \textbf{111.0} 
& 64.6 & 85.8 & 87.2  \\                               
                               & PSSA & 110.9 
& \textbf{74.1} & \textbf{96.3} & \textbf{93.8} \\

\midrule
\multirow{2}{*}{\makecell{Spike self- \\ attention }} & SDSA-1 & 104.3
& 54.8 & 47.1 & 68.7  \\
                               & SDSA-3 & 61.5 
& 56.1 & 30.7 & 49.4  \\
\midrule
\multirow{3}{*}{\makecell*[c]{Normalization \\ Method}} & None & 89.7
& 49.1 & 21.7 & 53.5  \\
                               & tdBN & 110.8 
& 64.2 & 78.5 & 84.5  \\
                               & tdLN & \textbf{111.0} 
& 72.0 & 93.1 & 92.0  \\
\bottomrule
\end{tabular}
\end{adjustbox}
\label{tab:Aablation}

\end{table}

\subsection{Ablation Study} 
We conduct ablation studies on spiking self-attention mechanisms and normalization method. More ablations including SNN timestep $T$ and local window size in PSSA are presented in \textbf{Appendix G}.

\paragraph{Design of Spiking Self-Attention Mechanisms}
To assess the impact of spike-driven self-attention mechanisms in offline RL tasks, we replaced our designs with those used in other SNN transformers, including SDSA-1 from \cite{yao2023spikedriven} and SDSA-3 from \cite{yao2024spikedriven}, while keeping all other components in DSFormer unchanged. Validation was conducted in the MuJoCo-Hopper environment. The results in Table~\ref{tab:Aablation} show that both TSSA and PSSA consistently outperform alternative mechanisms by a substantial margin. The performance limitation of SDSA-1 arises from its simplistic computation, which restricts the model's feature extraction capacity, making it less effective in complex problem modeling. For SDSA-3, its poor performance primarily because it was designed for vision tasks that emphasize feature-dimension relationships rather than token-to-token dependencies. Given that offline RL is an autoregressive task requiring next-token prediction based on previous states, SDSA-3 is less effective in this context.

\paragraph{Normalization Methods}
To evaluate the effectiveness of Progressive Threshold-dependent Batch Normalization (PTBN), we compared it against using no normalization, the commonly used tdBN in SNNs, and tdLN—a LayerNorm variant we proposed for sequence modeling. These experiments were conducted on DSFormer with PSSA in the MuJoCo-Hopper environment. As shown in Table~\ref{tab:Aablation}, performance significantly degrades without a Norm module, underscoring its importance in preventing gradient explosion and vanishing. PTBN achieves performance comparable to tdLN, while tdBN falls behind both. As discussed, tdBN is less effective for sequence modeling, and while tdLN offers stability, it disrupts SNN spiking properties. PTBN effectively combines the advantages of LayerNorm and BatchNorm, preserving temporal dependencies while maintaining the spiking nature essential to SNNs.

\begin{table}[t]
\centering
\caption{Results on AntMaze with varied sequence length}
\begin{adjustbox}{width=\columnwidth}
\begin{tabular}{ccccc}  
\toprule
AntMaze Mean & Length=50 & Length=100 & Length=150 & Length=200  \\
\midrule
\textbf{TSSA (ours)} & \textbf{65} & \textbf{73} & \textbf{71} & \textbf{72}  \\
DT & 62 & 67 & 58 & 59  \\ 
\bottomrule
\end{tabular}
\end{adjustbox}
\label{tab:long-range}
\end{table}

\begin{table}[t]
\centering
\caption{Results on D4RL-Hopper datasets with sparse reward}
\begin{adjustbox}{width=\columnwidth}
\begin{tabular}{c|ccc|ccc}  
\hline
\multirow{2}{*}{Task Name} & \multicolumn{3}{c|}{Dense Setting}  &
\multicolumn{3}{c}{\textbf{Sparse Setting}}  \\

& CQL & DT & DSFormer & CQL & DT & DSFormer \\
\hline
Hopper-m-e & 111.0 & 107.6 & 110.9 & 9.0 & 107.3 & \textbf{110.7} \\ 
Hopper-m & 58.0 & 67.6 & 74.1 & 5.2 & 60.7 & \textbf{68.2} \\
Hopper-m-r & 48.6 & 82.7 & 96.3 & 2.0 & 78.5 & \textbf{88.9} \\
\hline
\end{tabular}
\end{adjustbox}
\label{tab:sparce}

\end{table}

\subsection{Discussion}
In this section, we investigate the reasons of why DSFormer is effective in offline RL from the perspectives of long-range temporal dependencies and sparse reward processing.

\paragraph{Long-range Temporal Dependencies}
We evaluated DSFormer on tasks with substantial long-range temporal dependencies, specifically the AntMaze environment, which involves sparse rewards and complex maze navigation using an 8-DoF ``Ant" quadruped robot in a multi-goal, non-Markovian setting. To compare DSFormer with DT, we tested sequence lengths of 50, 100, 150, and 200 steps. As shown in Table~\ref{tab:long-range}, DSFormer consistently outperforms DT, with less performance degradation as sequence length increases. This advantage arises from the threshold-based activation in SNN neurons, which accumulate membrane potential and fire only when a threshold is met, then reset. This mechanism allows SNNs to remain inactive between critical events, preserving information over extended intervals without continuous updates, thus enhancing sensitivity to prior states and improving performance in tasks with significant temporal gaps.

\paragraph{Sparse Reward Processing}
To assess DSFormer’s advantage in sparse reward settings, we compared the performance of DSFormer, DT and CQL under sparse (delayed) reward conditions in the MuJoCo-Hopper tasks, where rewards are granted only at the final timestep of each trajectory. Table~\ref{tab:sparce} demonstrates that delayed returns minimally affect DSFormer and DT while CQL collapses. This resilience is due to SNNs’ event-driven nature, which triggers neuron activation only in response to key events, such as critical state changes or sparse reward signals, while remaining inactive during less relevant phases. This selective activation allows SNNs to focus on essential information, reducing interference from irrelevant data and enhancing responsiveness to reward-related events.

\section{Conclusion}

DSFormer is the first spike-driven transformer for offline RL via sequence modeling, employing novel self-attention mechanisms TSSA and PSSA to capture temporal and positional dependencies and PTBN normalization to replace SNN-incompatible LayerNorm while preserving temporal dependencies. The threshold-based activation in SNN neurons enables DSFormer to handle long-range dependencies and sparse environments effectively. Results on the D4RL benchmark indicate that DSFormer outperforms both SNN and ANN counterparts while achieving 78.4\% energy savings, demonstrating strong potential for energy-efficient embodied applications. A current limitation is the lack of neuromorphic chip deployment, which will be the focus of future work.

\paragraph{Acknowledgments}
This work is supported by Shanghai Artificial Intelligence Laboratory and the National Natural Science Foundation of China  under Grant No.62106139.

{\small
\bibliographystyle{ieee_fullname}
\bibliography{egbib}
}

\clearpage

\end{document}